\documentclass[letterpaper, 10 pt, conference]{ieeeconf}  
\usepackage{graphicx}
\usepackage{cite}
\usepackage{picinpar}
\usepackage{amsmath}
\usepackage{url}
\usepackage{flushend}
\usepackage{colortbl}
\usepackage{soul}
\usepackage{multirow}
\usepackage{pifont}
\usepackage{color}
\usepackage{alltt}
\usepackage{enumerate}
\usepackage{siunitx}
\usepackage{epstopdf}
\usepackage{pbox}
\usepackage{subcaption}

\IEEEoverridecommandlockouts                              

\usepackage{times}
\usepackage{epsfig}
\usepackage{graphicx}
\usepackage{amsmath}
\usepackage{amssymb}
\usepackage{cite}
\usepackage{subcaption}
\usepackage{algorithm}
\usepackage[utf8]{inputenc}
\usepackage{listings}
\usepackage{color}
\usepackage{authblk}
\definecolor{codegreen}{rgb}{0,0.6,0}
\definecolor{codegray}{rgb}{0.5,0.5,0.5}
\definecolor{codepurple}{rgb}{0.58,0,0.82}
\definecolor{backcolour}{rgb}{0.95,0.95,0.92}
\lstdefinestyle{mystyle}{
   backgroundcolor=\color{backcolour},
   commentstyle=\color{codegreen},
   keywordstyle=\color{magenta},
   numberstyle=\tiny\color{codegray},
   stringstyle=\color{codepurple},
   basicstyle=\footnotesize,
   breakatwhitespace=false,
   breaklines=true,
   captionpos=b,
   keepspaces=true,
   numbersep=5pt,
   showspaces=false,
   showstringspaces=false,
   showtabs=false,
   tabsize=2
}

\title{\LARGE \textbf {Object recognition and tracking using Haar-like Features Cascade Classifiers: Application to a quad-rotor UAV.}}

\author{Luis Arreola$^1$, Gesem Gudiño$^{1, 2}$ and Gerardo Flores$^1$
\thanks{$^1$Perception and Robotics Laboratory, Center for Research in Optics, Le\'{o}n, Guanajuato, Mexico, 37150.}
\thanks{$^2$Deparment of Industrial Electro-Mechanics, Universidad Tecnol\'{o}gica de Le\'{o}n,  Le\'{o}n, Guanajuato, Mexico, 37670.}
\thanks{(E-mail: luisfranciscoac@yahoo.com, gesemgudino@gmail.com, gflores@cio.mx). Corresponding author: Gerardo Flores.}
\thanks{This work was supported in part by the FORDECYT-CONACYT under grant 000000000292399 and the Laboratorio Nacional de \'Optica de la Visi\'on of the National Council of Science and Technology in Mexico (CONACYT) under agreement 293411.}
}
\begin{document}
\maketitle
\begin{abstract}
In this paper we develop a functional Unmanned Aerial Vehicle (UAV), capable of tracking an object using a Machine Learning-like vision system called Haar feature-based cascade classifier. The image processing is made on-board with a high processor single-board computer. Based on the detected object and its position, the quadrotor must track it in order to be in a centered position and in a safe distance to it. The object in question is a human face; the experiments were conducted in a two-step detection, searching first for the upper-body and then searching for the face inside of the human body detected area. Once the human face is detected the quadrotor must follow it automatically. Experiments were conducted which shows the effectiveness of our mythology; these results are showing in a video.
\end{abstract}
\section{Introduction}
\begin{figure}[t]\vspace{0.35cm}
  \centering
    \includegraphics[width=0.45\textwidth]{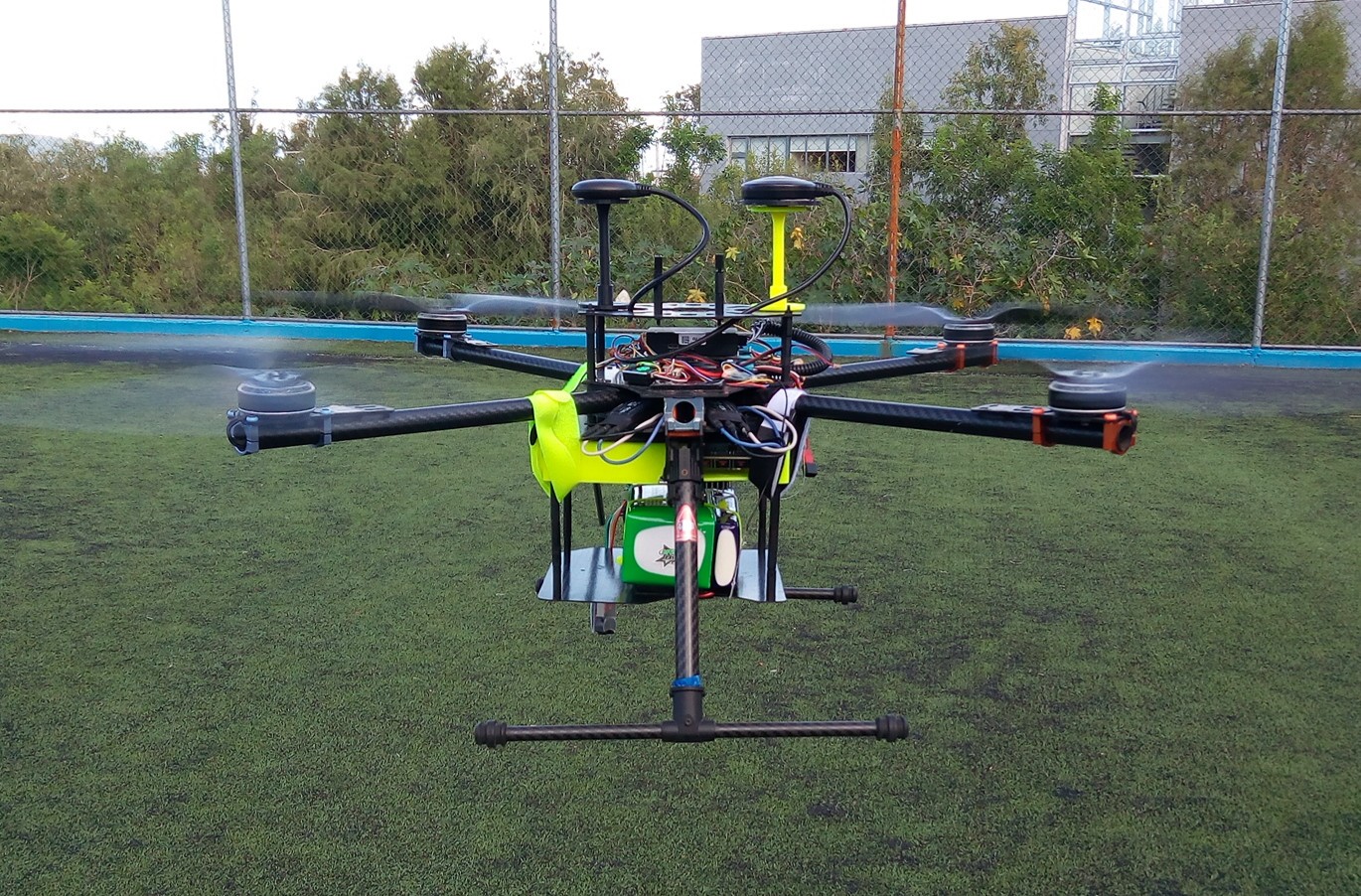}
  \caption{The quad-rotor unmanned aerial vehicle used in this study. It has a Jetson TX2 computer and a monocular camera on-board to perform the person detection.}
  \label{drone}
\end{figure}

The use of quadrotors in real applications such as aerial photography, environment monitoring, farming, structure inspection among others, demand implemented vision techniques to provide perception capabilities to the aerial drone \cite{andres}, \cite{rlozano}, \cite{aerial}, \cite{luis}. A camera-based system is one of the favorite solutions to these demands due to its passive and low-cost characteristics. This work presents an easy-to-implement application to this topic, programming object detection algorithms completely embedded in a quadrotor UAV which includes on-board computer and a monocular camera. The entire system consists in detecting an object by taking a frame from the camera and then the on-board computer processes the image to detect the object using a Haar-like feature-based classifier. Due to the fact that the Haar classifiers are considered as \textit{weak} classifiers \cite{weak}, a cascade training is implemented to obtain a robust detection. Once the object is detected, the on-board computer determines the position of the object with respect to (w.r.t) the quadrotor, hence it sends the corresponding information to the flight controller to ensure the correct tracking. For this problem, tracking refers to the capability that the quadrotor has to detect a desired object, in this case a person, and follow it with the information obtained by the vision algorithms. The UAV used for this approach can be seen at Fig. \ref{drone}.

There are several works that have explored object detection and tracking using UAVs \cite{lewis}, \cite{gaspar}. Next we cite several works related with our research. In \cite{zhang}, the object detection and tracking performed by an UAV is conducted with a video previously recorded. A face detection algorithm is presented in \cite{similar}, where image processing is embedded in an Odroid micro computer. Also the number of steps to determine the position of the object with respect to the UAV are reduced due to the simplicity of the performed strategies, which may result in a faster response from the UAV with respect to the persons movement. In \cite{jetsononbard} a Jetson TX2 is also implemented with object detection techniques in order to obtain better performance results.

A Vision-based navigation has developed multiple strategies depending on the applications, such as road following, power lines inspection, and navigation in orchards \cite{aouf}, \cite{stefas}, \cite{road}. For this work the general strategy consists in tracking a moving object, in particular a person's face, with the information of the continuous frames that are acquired by the camera. Each one of the frames are compared with the cascade training of the features classifiers to detect the person's face. The tracking consists in navigating the UAV in a way to achieve that the center of the image corresponds with the centroid of the object and with a certain predefined distance. This operating principle is the same used for vision-based autonomous landing, in which the camera must detect a preset landmark that indicates where the quadrotor should be land \cite{land1}, \cite{land2}. Several experiments were conducted in this work. As is briefly mentioned above, such experiments consist in detection and tracking of a human face. The detection process consists in searching for an upper body (from the shoulders to the head), and inside the sub-image of a \textit{detected upper-body} the human face is now searched withing such area. It is demonstrated trough experiments that this approach makes the system more robust than without the use of upper-body detection part.

The remainder of this paper is organized as follows. In Section \ref{sec:problem} the general approach and the architecture of the system are described. Section \ref{sec:exp} presents the results of the upper-body and face detection combined, and the subsequent tracking by the UAV by using the proposed approach. Finally, Section \ref{conclusion} presents some concluding remarks and an outline of future directions of the presented research.
\section{System description}\label{sec:problem}
In this section, the main hardware and software elements of the quadrotor tracking system are explained.
\subsection{Hardware}
\subsubsection{UAV}
Main elements of our quadrotor experimental platform includes: electronics control systems, a frame, Electronic Speed Controllers (ESCs), motors, propellers, a battery, a control board and an Inertial Measurement Unit (IMU) embedeed in an autopilot. The IMU is composed of a 3-axis accelerometer and a gyroscope. Its function is to get the actual information of the quadrotor's attitude. Also the quadrotor is endowed with a GPS module which gets the quadrotor's absolute position. The quad-rotor used in this paper is shown at Fig. \ref{drone}. This drone is equipped with all the primary elements listed above, including a Jetson TX2 single board computer and a voltage regulator. The quad-rotor specifications are listed in Table \ref{table:UAVdescription}.

\begin{table}[htb!]
\centering
\begin{tabular}{llr} 
\hline
\multicolumn{2}{c}{Quadrotor characteristics} \\
\cline{1-2}
Parameter & Value\\
\hline
Span      & 70 [cm]  \\
Height    & 26 [cm]       \\
Weight    & $\approx$ 2 [kg] \\
Propulsion& Brushless motor 330 [kv]    \\
        & Propeller 17x5.5 [in]\\
Max. Load & $\approx$ 3.5 [kg]\\

Battery type   & Li-Po 6s\\
            & Capacity 10000 [mAh]    \\
Flight controller & Pixhack v3 \\
Firmware  & ArduCopter 3.5.5 \\
Estimated flight time & 15 [min]\\
\hline
\end{tabular}
\caption{Quadrotor UAV parameters.}
\label{table:UAVdescription}
\end{table}

\subsubsection{NVIDIA Jetson TX2}
A Jetson TX2 developer kit is used to embed the vision and estimation algorithms. The training for the object detection is stored in the Jetson TX2. Depending on the object's position, the corresponding commands are send to the autopilot in order that UAV tracks the person's face. The specifications of this single-board computer can be seen at \cite{jetson}.

\subsection{Software}
The general code is based in two main parts. The first, describes all the training steps for the object detection task. The second part refers to the communication between the on-board computer and autopilot in order to perform control commands in the quadrotor. Then, when an object is detected, a bounding box surrounds it and calculates its centroid, then the navigation commands are computed as a function of the centroid position to achieve that the center of the image coincides with the centroid of the object. Once this step is done, the bounding box width determines if the UAV needs to make a pitch-forward to get closer to the object or not. If no object is detected, the UAV remains in its position in hover. When an action is performed, the program waits for the next frame and repeats the loop.

Object tracking stops in two cases: a) when the UAV's battery reaches a established minimum; and b) when the user decides to finish this task, for that, the user can turn off the tracking object remotely. Once the program ends, for any of the two reasons mentioned above, the quadrotor enters in \textit{failsafe} mode, which causes it to position itself at five meters high w.r.t takeoff position and once there, it returns to home position for landing.

\subsubsection{Haar cascades Training}
\begin{figure}[!htb]
  \centering
\includegraphics[width=0.48\textwidth]{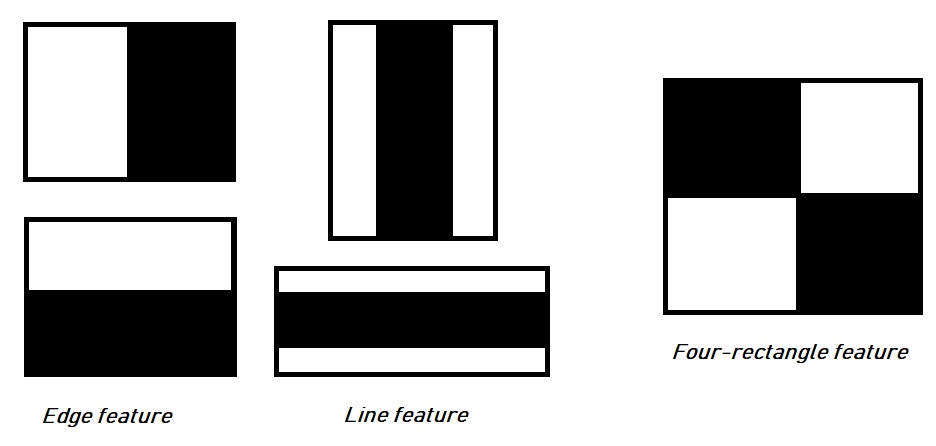}
  \caption{Different patterns considered as Haar features.}
  \label{haar_features}
\end{figure}

The Haar feature-based cascade classifier is an effective object detection method based on image information. It is a machine learning approach where a cascade function is trained from positive and negative images. When one says \textit{positive images} it means that the object of interest is in such image; the \textit{negative images} are the opposite, i.e., when the object of interest is not in the image. The object detection procedure classifies images based in the value of features, instead of working directly with pixels. The system uses three kind of features shown at Fig. \ref{haar_features}. 

%
The value of a \textit{two-rectangle feature} is the difference between the sum of the pixels within two regular regions (edge features). The regions must have the same size and shape, and are adjacent either horizontal or vertical. A \textit{three-rectangle feature} calculates the sum within two rectangles subtracted from the sum in a rectangle in the center (line feature). And finally a \textit{Four-rectangle feature} computes the difference between two pairs of rectangles diagonally positioned. Given a base resolution of the detector as $24x24$ pixels, it results in a large number of rectangle features, over $180,000$. In order to compute this features rapidly, there exists a technique called \textit{integral image}. The integral image at a given location $(x,y)$ in pixels contains the sum of the pixels left and above of $(x,y)$ and it is given by the following formula
\begin{equation}
ii(x,y) = \sum_{x'\leq x,y'\leq y}i(x',y')
\end{equation}
where $ii(x,y)$ is the integral image at location $(x,y)$ and $i(x',y')$ is the original image at location $(x,y)$ \cite{viola}. The integral image can be computed following the next pair of recurrences
\begin{equation}
\centering
\begin{aligned}
&s(x,y) = s(x,y-1) + i(x,y)\\
&ii(x,y) = ii(x-1,y) + s(x,y)
\end{aligned}
\end{equation}
where $s(x,y)$ is the cumulative sum of a row, therefore $s(x,-1)$ = 0 and $ii(-1,y) = 0$.

\begin{figure}
\centering
\begin{subfigure}[b]{0.45\textwidth}
\includegraphics[width=\textwidth,height=3.5cm]{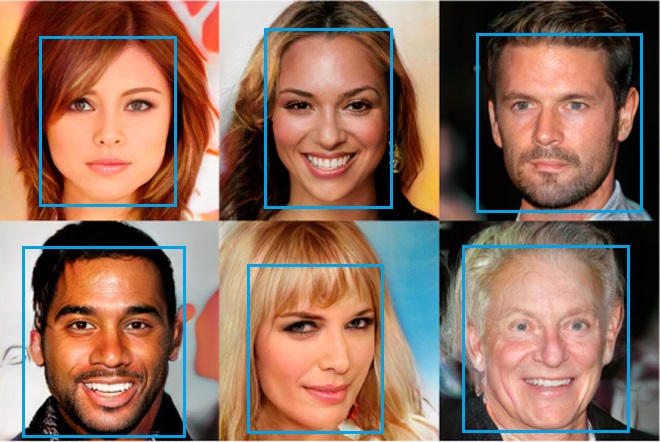}
\caption{Positive images.}
\label{fig:caras}
\end{subfigure}
\begin{subfigure}[b]{0.45\textwidth}
\includegraphics[width=\textwidth,height=3.5cm]{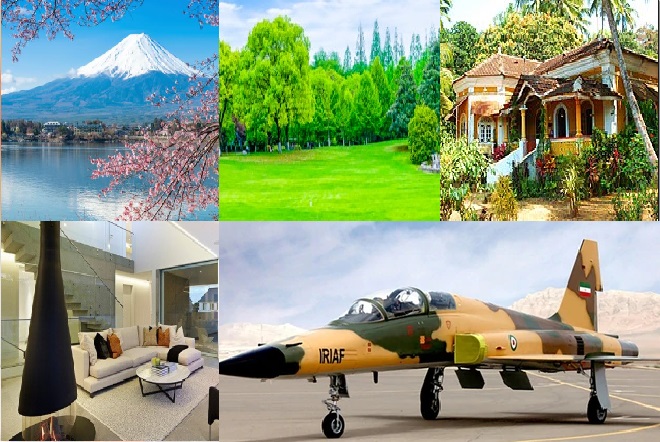}
\caption{Negative images.}
\label{fig:nocaras}
\end{subfigure}
\caption{Example of positive and negative set of images. In the positive set of images it is shown a bounding box surrounding the object of interest.}\label{set_images}
\end{figure}

Initially, the algorithm needs a big number of positive images and negative images. The recommended number of images is of the order of a few thousand of images for the positive set and five hundred images for the negative set \cite{opencv1}, \cite{opencv2}. An example of positives and negatives sets of images is shown at Fig. \ref{set_images}
\begin{enumerate}
    \item \textbf{Negative samples.}
    Negative samples are the set of images where the object of interest, in this case a person's face, can not be found, in other words, it contains everything the user does not want to detect. The set of negative images are taken from arbitrary images and must be prepared by the user manually  and are enumerated in an text file. In such text file it is described the image file name and  directory information. An example of negative samples is shown at Fig. \ref{fig:nocaras}. The images in this set can be of different sizes, but each image should be at least equal or larger than the desired training windows size. The set of negative window samples will be used to tell the features classifier what not to look for when trying to find the object of interest.
    \item \textbf{Positive samples.}
    There exist two possible ways to generate the positive samples. One of them is making use of an OpenCV library function called \texttt{\$opencv\_createssamples}. This method boosts the process to define what the model should look like generating artificial images with the desired object. The second method is selecting manually the positive images. In this project the second method was chosen. Like with the negative samples, the user manually prepares the set of images. To ensure a robust model, the samples should cover a wide range of varieties that can occur within the object class. In the case of faces, the samples must consider different gender, emotions, races and even beard styles. An example of positive samples is shown at Fig. \ref{fig:caras}. For the upper body, the samples must consider different positions, haircuts styles and sizes. Inside the directory where the images are located, a \texttt{.dat} file needs to be included. Each line of this file corresponds to an image, the first element of the line is the name of the image, followed by the number of objects within the image, followed by numbers indicating the coordinates of the object(s) bounding rectangle(s), i.e. $(x, y, $width$, $height$)$ \cite{opencv3}.
\end{enumerate}
    
\begin{figure}[!htb]
\centering
  \includegraphics[width=0.5\textwidth]{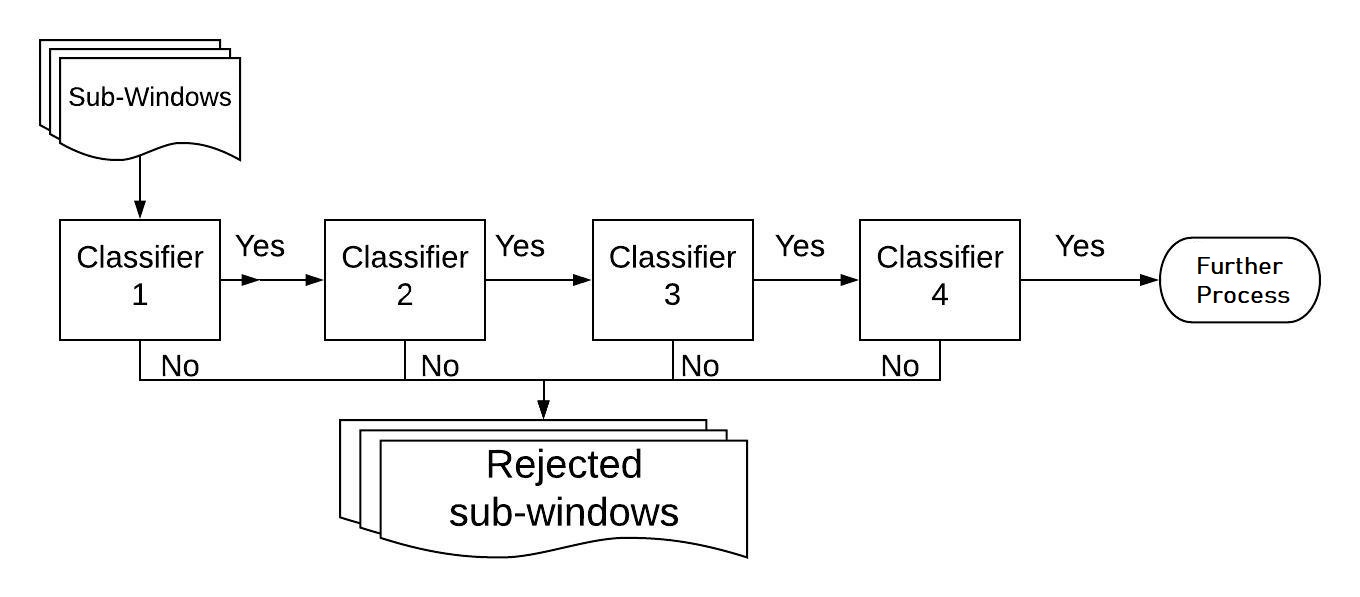}
    \caption{Schematic description of a detection cascade.}
    \label{cascade}
\end{figure}

The cascade classifier is an algorithm that achieves an improved detection performance while reducing the computation time. The overall form of a cascade classifier is that of a degeneration tree. A positive result from one first classifier leads to a evaluation of a second classifier that has been adjusted to achieve high detection rates. If the second classifier gives a positive result, it triggers a third classifier, and so on. In any stage, if a negative result is thrown by the algorithm, then an immediate rejection of the sub-window is performed \cite{pviola}. This is shown in a graphic way in Fig. \ref{cascade}.
\\
\textbf{Training Haar cascades.}
Once the two sets of images are prepared, the Haar cascade training is ready to be performed. Using OpenCV, the command to run is \texttt{opencv\_traincascade}, that looks like:
\texttt{\$ opencv\_traincascade -data data -vec positives.vec -bg bg.txt -numPos 1800 -numNeg 900 -numStages 10 -w 20 -h 20}
Where \texttt{-data} refers to the direction where the trained classifier should be stored; \texttt{-vec} refers to the vec-file with positive samples; \texttt{-bg} refers to the background description file. This is the file direction containing the negative sample images; \texttt{-numPos} refers to the number of positive samples used in the training for every stage; \texttt{-numNeg} refers to the number of negative samples used in training for every stage; \texttt{-numStages} refers to the number of cascade stages to be trained. The more stages are selected, the more robust the training will be, however, it can take a longer time to compute depending on the training computer capacities; and finally \texttt{-w} and \texttt{-h} refer to the width and height of the training samples in pixels \cite{haar}. The output files are going to depend of the number of stages that the user selected. In the example, the directory \texttt{data} contains eleven \texttt{.xml} cascade training files, one per stage; and also contains the \texttt{final.xml} file that represents the whole cascade training.

\subsubsection{Dronekit-Python}
Dronekit-Python is a tool that runs in an on-board computer and allows users to create a communication between UAV's on-board computers and the ArduPilot flight controller, all these using a low latency link. The on-board processing enhances the autopilot efficiency significantly, improving the behavior of the vehicle and performing tasks that are computationally intensive or time-sensitive, such as computer vision and estimation algorithms or even path planning \cite{dronekit}. The main use of Dronekit-Python in this paper, consists in arming the data package that contains the information about the velocity and directions that the quadrotor needs to perform given the information of the object detection. These data packages connect with the flight controller using MAVLink, a communication protocol for small UAV's \cite{MAVLink}, \cite{MAVLink2}.

\begin{figure}[h]
\centering
\includegraphics[width=0.5\textwidth,height=5cm]{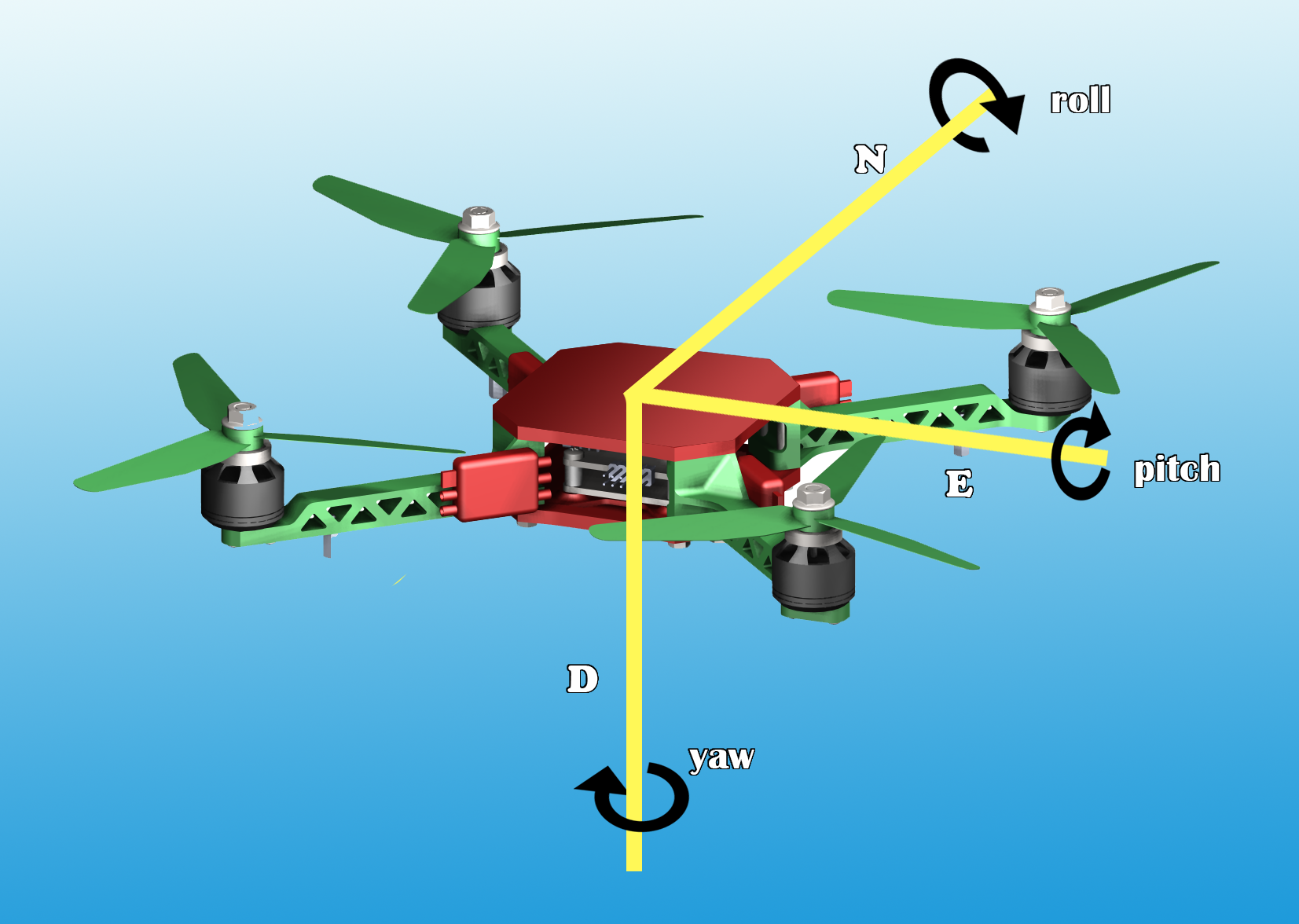}
\caption{NED axes with respect to the body's current frame.}
\label{dron_NED}
\end{figure}

The Dronekit function \texttt{Send\_NED\_Velocity} is used for making a connection between Jetson TX2 and the PixHack-V3 flight controller, the autopilot used in this work. The function is defined as follows
\begin{lstlisting}[language=Python, firstline=9, lastline=20]
def Send_NED_Velocity(velocity_x,velocity_y,velocity_z,duration,vehicle):
    msg = vehicle.message_factory.set_position_target_local_ned_encode(
        0,       # time_boot_ms (not used)
        0, 0,    # target system, target component
        mavutil.mavlink.MAV_FRAME_BODY_OFFSET_NED, # 
        0b0000111111000111, # type_mask
        0, 0, 0, # x, y, z positions (not used)
        velocity_x, velocity_y, velocity_z, # m/s
        0, 0, 0, # x, y, z acceleration
        0, 0)
    vehicle.send_mavlink(msg)
    #time.sleep(0.25)
\end{lstlisting}
The above mentioned function asks for several parameters, being \texttt{velocity\_x, velocity\_y, velocity\_z} and \texttt5{vehicle} the necessary parameters for our application. The first three parameters represent the velocities in North-East-Down (NED) directions, being in this case the pitch-forward, roll to the right and a negative thrust. The reasons for this, is that the directions are with respect to the body frame. Then it is necessary to specify the UAV position w.r.t the body frame, for that, we use the line \texttt{MAV\_FRAME\_BODY\_OFFSET\_NED}. The body frame is depicted at Fig. \ref{dron_NED}. Thanks to these parameter it can be specified the position in $x$ meters north, $y$ meters east and $z$ meters down of the current UAV position. Velocity directions are in the NED frame \cite{NED}.

%
\section{Experiments and Results}\label{sec:exp}
In this section the experiments and results obtained from the methods shown in section \ref{sec:problem} are described, emphasizing the efficiency in the object detection and object tracking performed autonomously by the UAV.
\subsection{Object Tracking}
\begin{figure}[h!]
\centering
\includegraphics[width=0.5\textwidth]{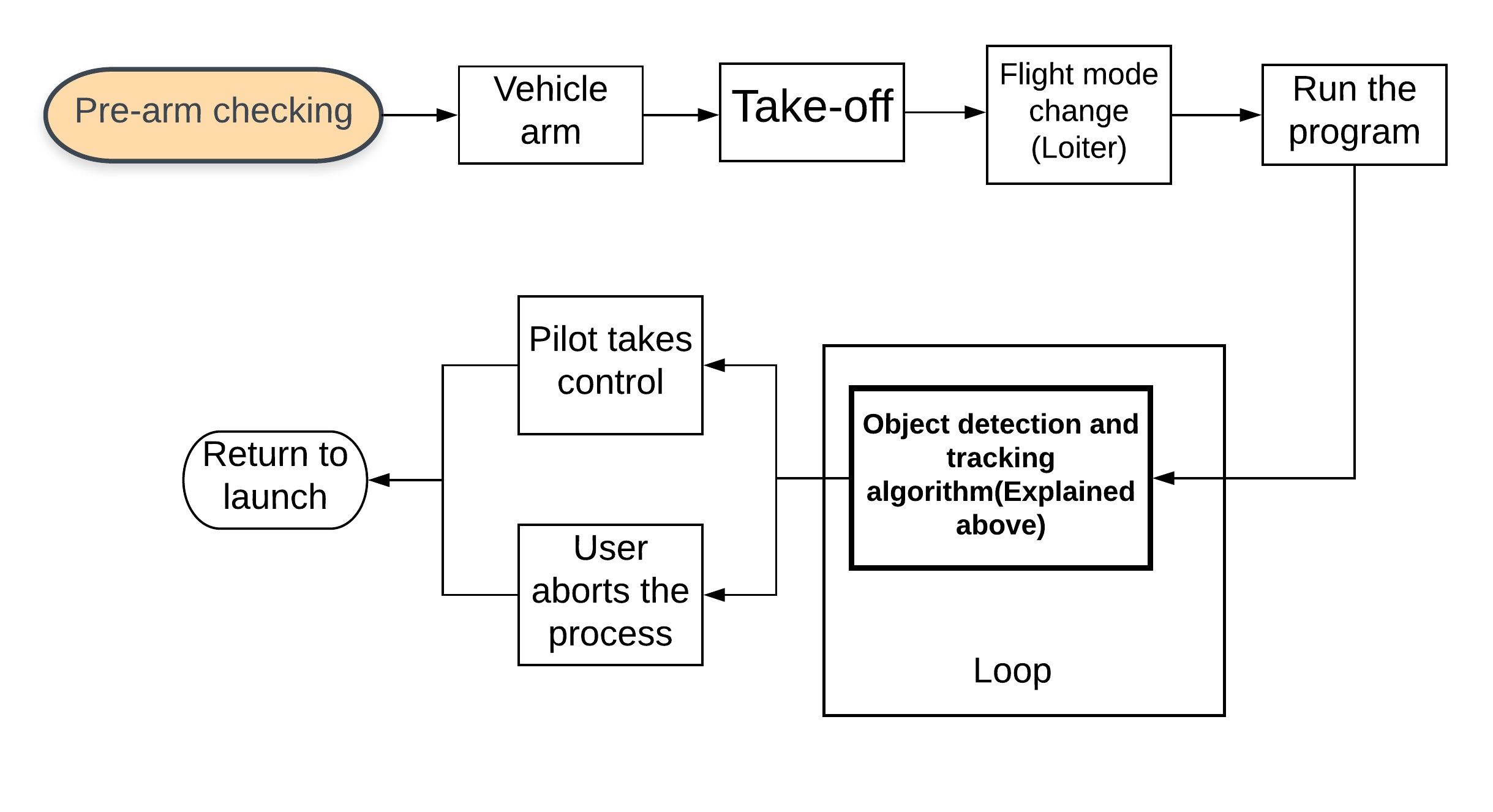}
\caption{Flow chart describing the experiments execution steps.}
\label{chart}
\end{figure}

The flowchart that describes the steps for the experiments in the rest of this section is shown at Fig. \ref{chart}.
\subsubsection{Object detection}\label{object}
The object to be detected for this experiment is a human face. We have chosen a human face due to the number of specific parts of a face that can be implemented in the training stage. These parts: eye, mouth, nose, etc. make the recognition more robust. In this case, the process to detect a face consists in two separate training processes:
\begin{enumerate}
\item \textbf{Upper body detection:} The Haar Feature Classifier algorithm is trained to detect the upper body of a human being. A rectangle is drawn covering the full area of the image where the upper body is detected, as it is shown in Fig. \ref{eye}. This training is not enough, because no matter how well the process of training is performed, the object detection sometimes may lead to detect false objects.
\begin{figure}[h!]
\centering
  \includegraphics[width=7cm,height=4cm]{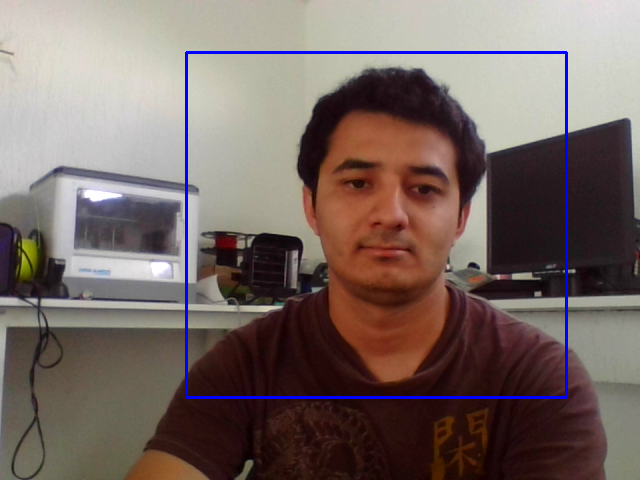}
    \caption{Upper body detection using Haar-cascade training.}
    \label{eye}
\end{figure}
\item \textbf{Complete face detection:} After the upper body detection, the Haar Feature Classifier algorithm is trained to detect a complete face. In this way, the detection goes from something general to a more specific object, in this case, the face. Like the previous case, rectangles are drawn around the detected object, as it is shown in Fig. \ref{face}.
\begin{figure}[h!]
\centering
  \includegraphics[width=7cm,height=4cm]{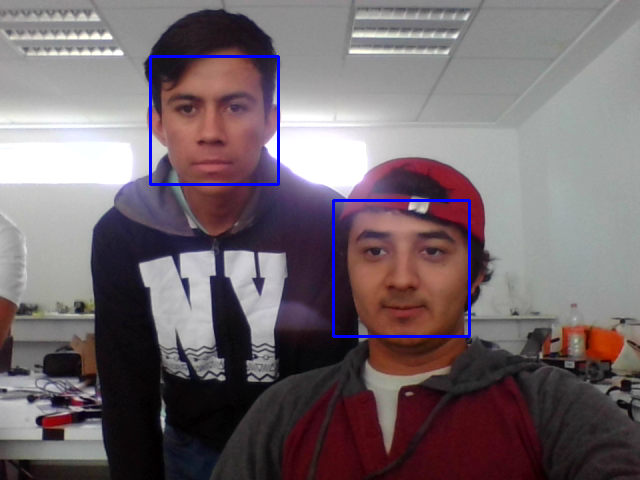}
    \caption{Face detection using Haar-Cascade training.}
    \label{face}
\end{figure}
\item \textbf{Combining both training stages:} As it has been mentioned before, no matter how good the training is, there always be a probability that a false object will be detected. To minimize this probability, the face detection training is made only inside a loop where the upper body detection is performed, as it is shown at Fig. \ref{full}. This methodology is conducted to ensure that no faces will be detected outside of an upper body and no face will be detected without an upper body, then the probability of detecting false objects is reduced.
\begin{figure}[h!]
\centering
\includegraphics[width=7cm,height=4cm]{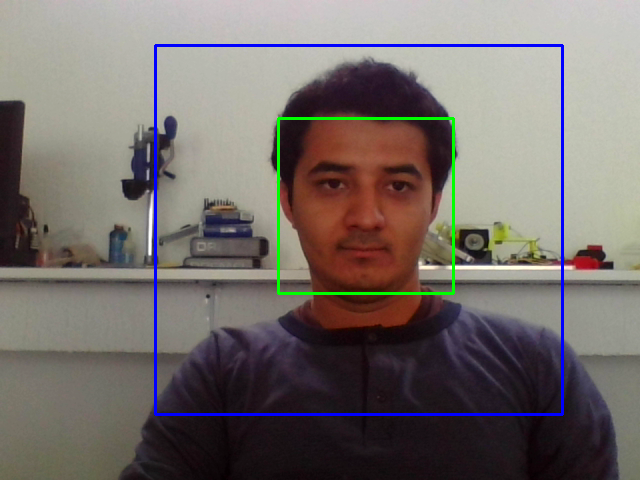}
\caption{Combination of upper body and face detection using Haar-cascade training.}
\label{full}
\end{figure}
\end{enumerate}
\subsubsection{UAV displacements}
\begin{figure}[t!]
\centering
\begin{subfigure}[h]{0.4\textwidth}
\includegraphics[width=\textwidth]{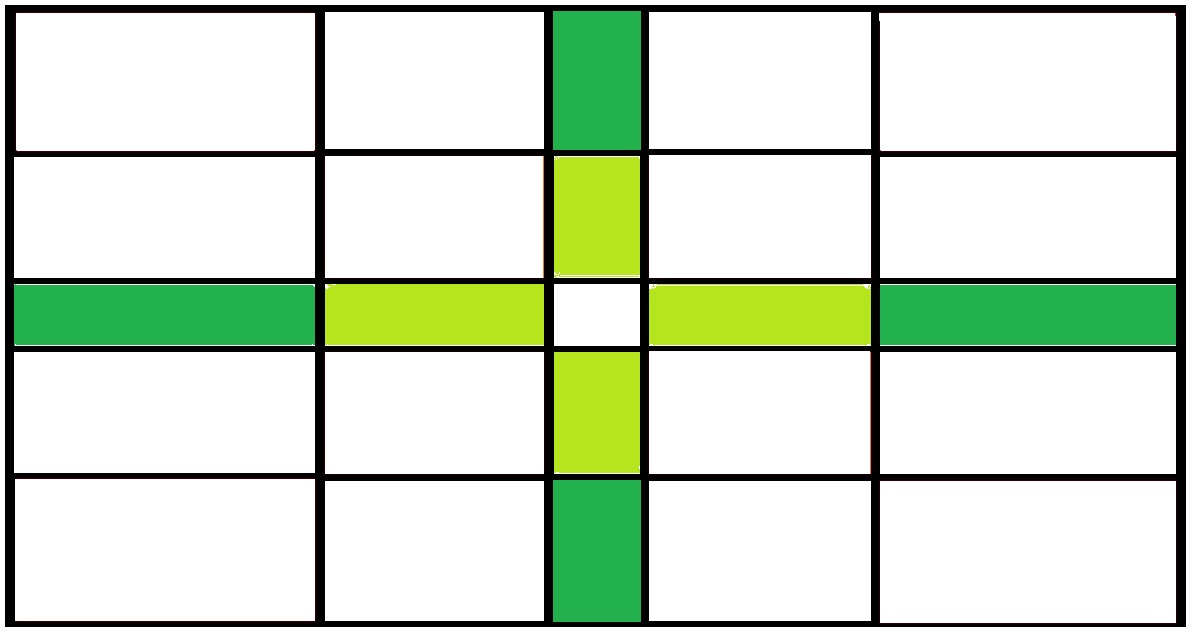}
\caption{Movements in simple direction. Soft green means slow movement, while hard green means fast movement.}
\label{fig:simpledir}
\end{subfigure}
\vspace{0.3cm}
\begin{subfigure}[h]{0.4\textwidth}
\includegraphics[width=\textwidth]{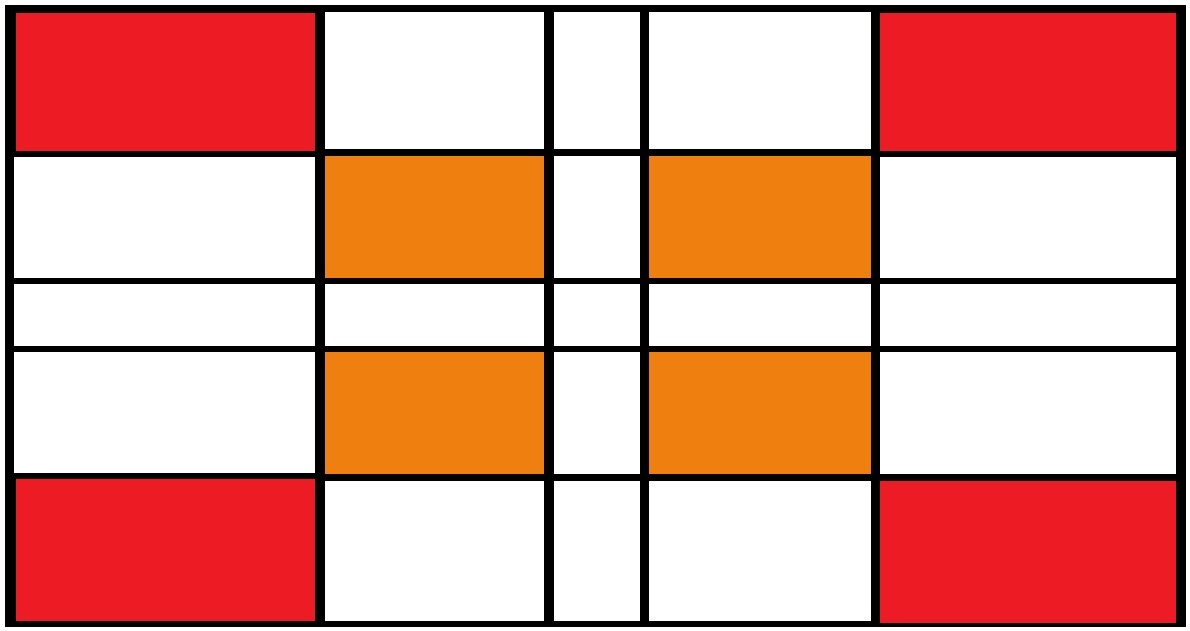}
\caption{Movements in two directions at same speed. Orange means slow movement, while red means fast movement.}
\label{fig:diagonal}
\end{subfigure}
\vspace{0.3cm}
\begin{subfigure}[h]{0.4\textwidth}
\includegraphics[width=\textwidth]{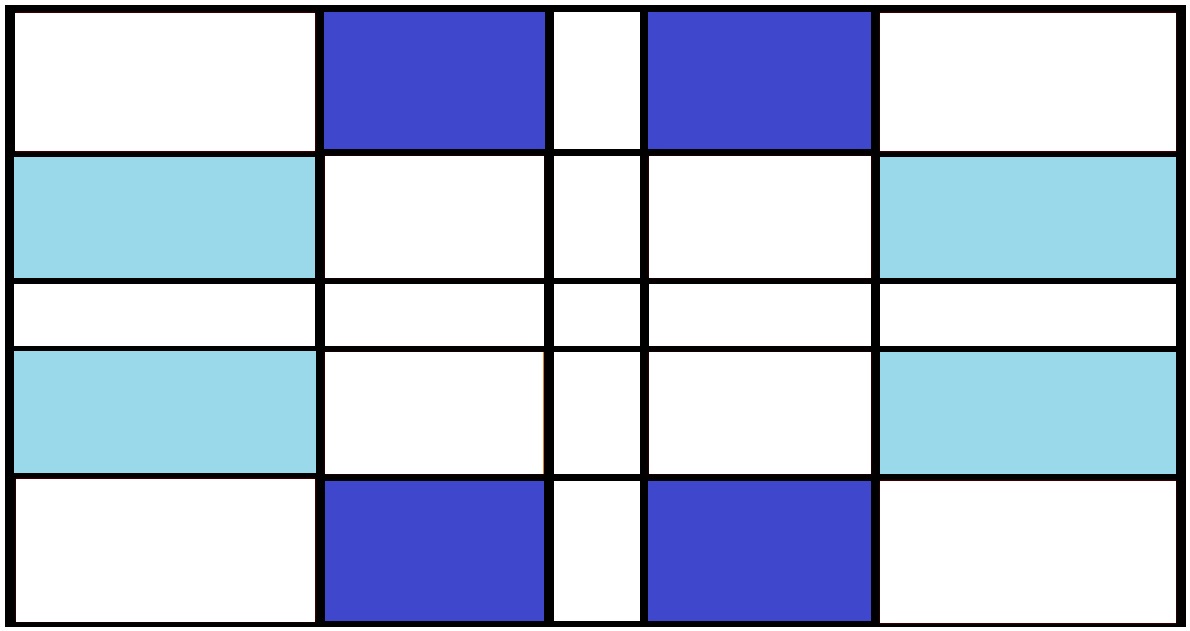}
\caption{Movements in two directions at different velocities. Soft blue means fast roll movement and slow thrust movement. Hard blue means fast thrust movement and slow roll movement.}
\label{fig:mix}
\end{subfigure}
\caption{Roll and thrust movements depending of the image zone where the centroid's object is detected.}\label{fig:zones}
\end{figure}

The next part of the experiment consists in identifying the the centroid's object which is located in the given image. When the centroid is not in the image center, it can be in one of three different zones, and depending in which one of these is, it will be the movement that the UAV will perform, these movements can also be classified in three types as shown in Fig. \ref{fig:zones}.

The quadrotor velocity and directions are assigned depending on the centroid position, as it is shown in the nex pseudo code:
\begin{lstlisting}[language=Python, firstline=57, lastline=69]
      elif x<290 and y<210:
         if x<145 and y<105:
            print ("Fast Left Up Fast")
            Send_NED_Velocity(0,-roll_f,-th_f,1,vehicle)
         elif x<145 and y>105:
            print ("Fast Left Down Slow")
            Send_NED_Velocity(0,-roll_f,-th_s,1,vehicle)
         elif x>145 and y<105:
            print ("Slow Left Up Fast")
            Send_NED_Velocity(0,-roll_s,-th_f,1,vehicle)
         elif x>145 and y>105:
            print ("Slow Left Up Slow")
            Send_NED_Velocity(0,-roll_s,-th_s,1,vehicle)
\end{lstlisting}

In this snipped code there are two main conditions that need to be satisfied for each one of the cases shown in Fig.\ref{fig:zones}. The first condition is to know where the objects centroid is w.r.t. the center of the image frame (top, bottom, right, left or any corner). The second condition is looking if the centroid location is actually near or far from the image center, making a decision to perform a slight or aggressive maneuver and roll displacement. For example, if the object centroid is located at the image top-left corner, the quadrotor will perform an aggressive roll and diminish thrust, as the object gets closer to the center it will decrease the velocity of the movements until it reaches the very image center. Variables that control these movements in the code are \texttt{roll\_f} and \texttt{roll\_s} for fast and slow movements in the roll angle, respectively, and \texttt{th\_f}, \texttt{th\_s} to define a fast and slow motors speed, respectively.
\begin{figure}
\centering
\begin{subfigure}[b]{0.49\textwidth}
\includegraphics[width=\textwidth,height=3.5cm]{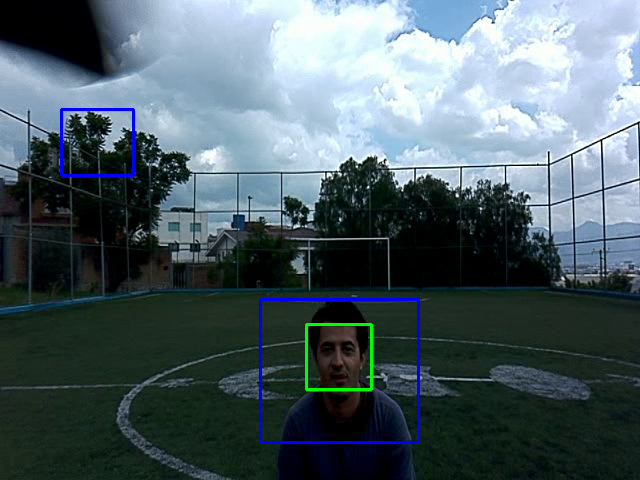}
\caption{Drone camera view 1.}
\label{dronecam5}
\end{subfigure}
\begin{subfigure}[b]{0.49\textwidth}
\includegraphics[width=\textwidth,height=3.5cm]{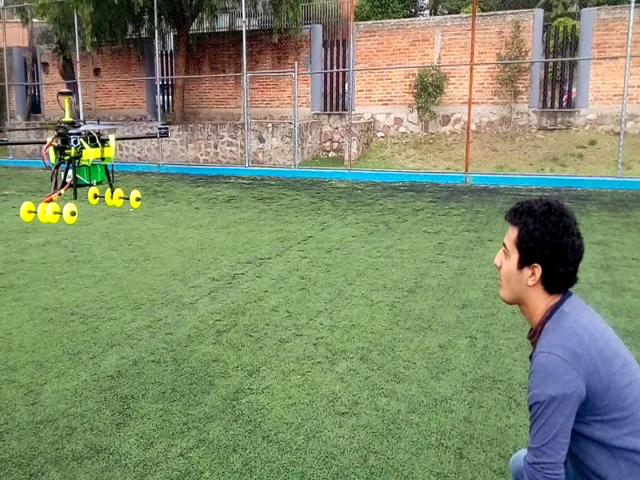}
\caption{External camera view 1.}
\label{extview5}
\end{subfigure}
\vspace{0.5cm}
\begin{subfigure}[b]{0.49\textwidth}
\includegraphics[width=\textwidth,height=3.5cm]{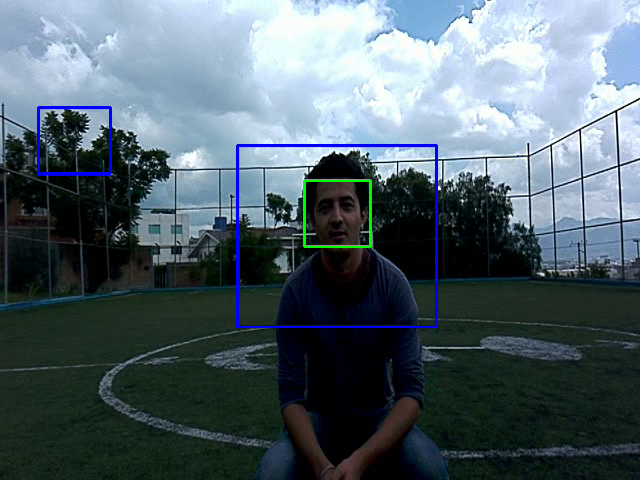}
\caption{Drone camera view 2.}
\label{dronecam6}
\end{subfigure}
\begin{subfigure}[b]{0.49\textwidth}
\includegraphics[width=\textwidth,height=3.5cm]{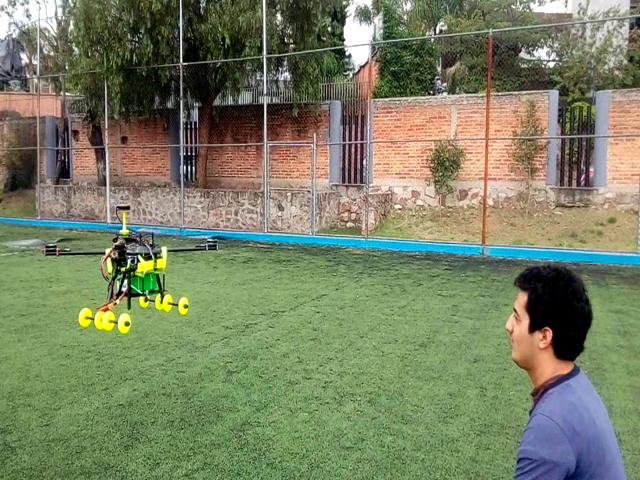}
\caption{External camera view 2.}
\label{extview6}
\end{subfigure}
\caption{Sequence of ordered object tracking images taken from the drone perspective and from an external video recording.}\label{fig:seq1}
\end{figure}

Fig. \ref{fig:seq1} shows two sequences of images representing the object recognition and tracking performed by the UAV. As it was stated in Section \ref{sec:problem}, there are some frames where false upper bodies are detected, but the real face is only recognized when a previous upper body is detected. The full video of the external view and the drone camera view can be found following the next link: \newline
\url{https://youtu.be/SY-dss_jJA4}

The video of the flight experiments shows how the face is detected and how the drone navigates following the face. The program analyzes each image with a rate of about four frames per second, having a reaction time between the object detection and the UAV movement of about $0.2$ seconds. It also can be observed that bounding boxes appear in places where the algorithm detects false upper bodies, but in none of these cases a face is found.

\section{Conclusions}\label{conclusion}
The object detection using the Haar Feature-based Cascade Classifier method seems to work relatively well for vision-based drone navigation. To ensure that the detection be reliable, the process must be meticulous and follow all the steps correctly. For example, at least $500$ positive images $250$ negative images are required to guarantee a good training. The decision to perform a two-step detection, one being the upper body and another the face, demonstrates better results in comparison with only face detection. The possibility to detect faces in a false upper body decreases exponentially, in fact, during the experiments there never is a case where a complete object, upper body + face, is detected wrongly.

Future works include: a) the capability of the UAV to perform yaw movements in case the object makes a rotation on its own axis. The most effective method could be detecting when the two points that form the horizontal top line of the bounding box are not in the same row of the image or in a certain hysteresis; b) using different methods for object detection including deep learning techniques; methods like convolutional neuronal networks should be addressed to make a comparison about their performance, particularly, TensorFlow can be implemented taking advantage of GPU features, therefore the Jetson TX2 would be better exploited.

\bibliographystyle{IEEEtran}
\bibliography{References}
\end{document}